# Extraction of Facial Feature Points Using Cumulative Histogram

**Sushil Kumar Paul [1], Mohammad Shorif Uddin [2] and Saida Bouakaz [3]**

**[1] Department of Computer Science and Engineering (CSE)**
**Jahangirnagar University, Savar, Dhaka-1342, Bangladesh**
**Phone: + (880) 1711172191**
E-Mail: **paulksushil@yahoo.com**

**[2] Department of Computer Science and Engineering (CSE)**
**Jahangirnagar University, Savar, Dhaka-1342, Bangladesh**
**Phone: +(880) 1552471751**
E-Mail: **shorifuddin@gmail.com**

**[3] Head of the SAARA Research Team, LIRIS Lab, Nautibus Building**
**University Claude Bernard Lyon1, 69622 Villeurbanne Cedex, France**
**Phone: +33 4 72 44 48 83**
**Fax: +33 4 72 43 13 12**
E-Mail: **saida.bouakaz@liris.cnrs.fr**

**Abstract**
This paper proposes a novel adaptive algorithm to extract facial feature points automatically such as eyebrows corners, eyes corners, nostrils, nose tip, and mouth corners in frontal view faces, which is based on cumulative histogram approach by varying different threshold values. At first, the method adopts the Viola-Jones face detector to detect the location of face and also crops the face region in an image. From the concept of the human face structure, the six relevant regions such as right eyebrow, left eyebrow, right eye, left eye, nose, and mouth areas are cropped in a face image. Then the histogram of each cropped relevant region is computed and its cumulative histogram value is employed by varying different threshold values to create a new filtering image in an adaptive way. The connected component of interested area for each relevant filtering image is indicated our respective feature region. A simple linear search algorithm for eyebrows, eyes and mouth filtering images and contour algorithm for nose filtering image are applied to extract our desired corner points automatically. The method was tested on a large BioID frontal face database in different illuminations, expressions and lighting conditions and the experimental results have achieved average success rates of 95.27%.

## 1. Introduction

Face analysis such as facial features extraction and face recognition is one of the most flourishing areas in computer vision like identification, authentication, security, surveillance system, human-computer interaction, psychology and so on [1]. Facial features extraction is the initial stage for face recognition in the field of vision technology. The most significant feature points are eyebrows corners, eyes corners, nostrils, nose tip, and mouth corners. These are the key components for face recognition [2], [3]. Eyes are the most crucial facial feature for face analysis because of its inter-ocular distance, which is constant among people and unaffected by moustache or beard [3]. Eyebrows, eyes and mouth also convey facial expressions. Another valuable face feature points are nostrils because nose tip is the symmetry point of both right and left side face regions and nose indicates the head pose and it is not impacted by facial expressions [4]. Therefore, face recognition is distinctly influenced by these feature points.

Currently, Active Shape Model (ASM) and Active Appearance Model (AAM) are extensively used for face alignment and tracking [5]. Facial feature extraction methods could be divided in two categories: texture-based and shape-based methods.

Texture-based methods take local texture e.g. pixel values around a given specific feature point instead of concerning all facial feature points as a shape (shape-based methods). Some texture-based facial feature extraction algorithms are: hierarchical 2-level wavelet networks for facial feature localization [6], facial point detection using log Gabor wavelet networks by employing geometry cross-ratios relationships[7], neural-network-based eye-feature detector by locating micro-features instead of entire eyes[8]. Some shape-based facial feature extraction algorithms including AAM, based on face detectors are: view-based active wavelet network [9], view-based direct appearance models [10]. The combination of texture- and shape-based algorithms are: elastic bunch graph matching [11], AdaBoost with Shape Constrains [12], 3D Shape Constraint using Probabilistic-like Output [13]. Wiskott et al. [11] represented faces by a rectangular graph which is based on Gabor wavelet transform and each node labelled with a set of complex Gabor wavelet coefficients. Cristinacce and Cootes [12] used the Haar features based AdaBoost classifier combined with the statistical shape model. In both ASM and AAM, a model is built for predefined points by using the test images and then an iterative scheme is applied to this model in detecting feature points. Most of the above mentioned algorithms are not entirely reliable due to variation in pose, illumination, facial expression, and lighting condition and high computational complexity. So, it is indispensable to develop robust, automatic, and accurate facial feature point localization algorithms, which are capable in coping different imaging conditions.

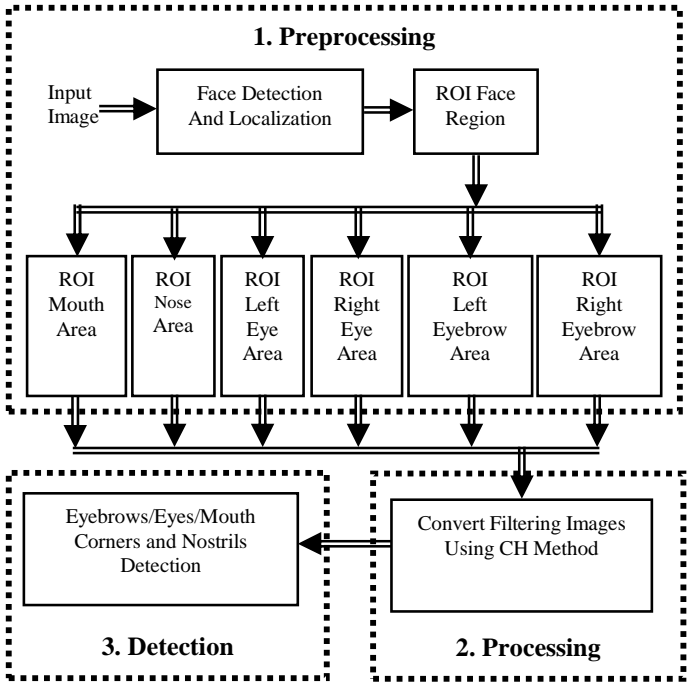

Figure 1. Block diagram of proposed feature extraction algorithm.

In this paper, we propose a robust adaptive algorithm based on cumulative histogram (CH) scheme that extracts the facial feature points in a fast as well as accurate way under varying illuminations, expressions and lighting conditions. Figure 1 shows the block diagram of our proposed algorithm that includes preprocessing, main processing and detection blocks. The preprocessing block detects the face and crops the face, right eyebrow, left eyebrow, right eye, left eye, nose, and mouth areas. The processing block is responsible for six ROIs such as right eyebrow, left eyebrow, right eye, left eye, nose, and mouth areas and then converts into binary images. The detection block detects the corner points of six ROIs. The remainder of the paper is organized as follows. Section 2 describes the region of interest (ROI) detection. In section 3, we present the mathematical description of CH method, which form the basis for our approach, and then we explain the facial feature point detection with the algorithmic details. Section 4 shows the experimental results of our facial feature extraction system. Finally we conclude the paper along with highlighting future work directions in section 5.

## 2 Region of Interest Detection

A rectangular portion of an image to perform some other operation and also to reduce the computational cost for further processing is known as region of interest (ROI).

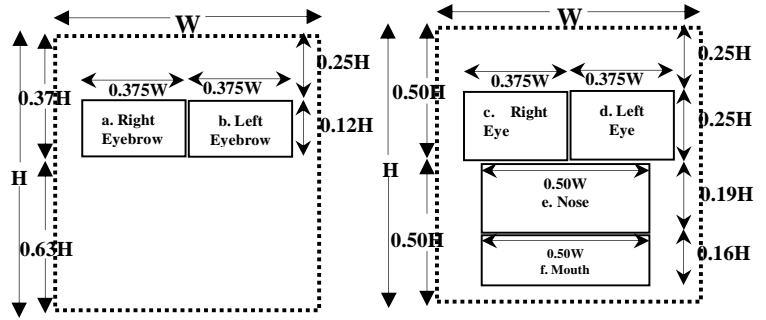

Figure 2. Location and size of six ROIs of a face image such as (a.) Right Eyebrow (Size:0.375W×0.12H), (b.) Left Eyebrow (Size:0.375W×0.12H), (c.) Right Eye (Size:0.375W×0.25H, (d.) Left Eye (Size:0.375W×0.25H,(e.) Nose (Size: 0.50W×0.19H), and (f.) Mouth (Size: 0.50W×0.16H) where, W=Image Width and H=Image Height.

By applying the Viola-Jones face detector algorithm, the detected face region is cropped first then we divide the face area vertically into upper, middle and lower parts [14]. From the human frontal face structure concept, eyebrows & eyes, nose, and mouth areas are situated in upper, middle, and lower portions of the face image, respectively. Again, the upper portion is partitioned horizontally into left and right segments for isolating right-eyebrow and right-eye and also left-eyebrow and left-eye, respectively.

Finally, the smallest ROI regions are segmented for right-eyebrow, right-eye, left-eyebrow, left-eye, nose, and mouth in order to increase the detection rate. Figure 1, Figure 2, and Figure 3(d) are shown the block diagram of our proposed algorithm, location and size of six ROIs and cropped images, respectively.

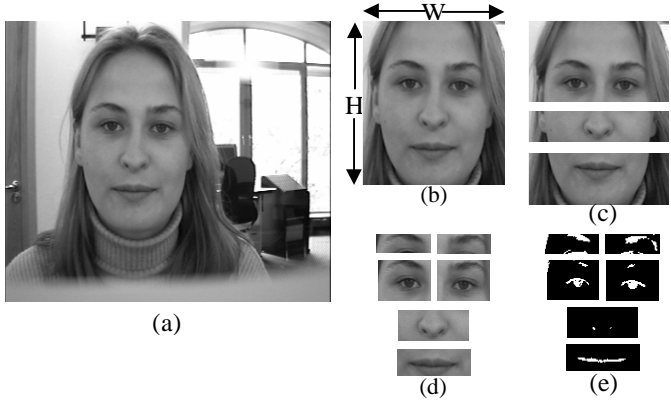

(a) (b) (c) (d) (e)

Figure 3. Procedure of our proposed algorithm: (a) Input image, (b) Detected and cropped the face, (c) Face is divided into three vertical parts, which are indicated eyebrows, eyes, nose and mouth areas, (d) Six ROIs show the exact right-eyebrow, right-eye, left-eyebrow, left-eye, nose and mouth regions, (e) Applying CH method, all of the six ROIs are converted into new filtering images.

## 3. Facial Features Extraction

Our proposed method exhibits the location of twelve crucial feature points including eight corner points for right-eyebrow, right-eye, left-eyebrow, left-eye, two points for nostrils and two corner points for mouth as shown in the Figure 3(d) and Figure 3(e). Feature points are extracted by an adaptive approach. To create the new filtering (binary) images, the following mathematical concepts are applied on each of the six original cropped (ROIs) gray scale images such as right-eyebrow, right-eye, left-eyebrow, left-eye, nose and mouth regions(see Figure 3(d) and Figure 3(e))[16],[17].

$$P_{I(x,y)}(v) = P(I(x,y)=v) = \frac{n_v}{N} \quad \text{Where,} \quad 0 \leq v \leq 255 \quad (1)$$

$$CH_{I(x,y)}(v) = \sum_{i=0}^{V} P_{I(x,y)}(i) \quad (2)$$

$$I_{FI}(x,y) = \begin{cases} 255 & \text{when} \quad CH(I(x,y)) \leq Th \\ 0 & \text{otherwise} \end{cases} \quad (3)$$

Where, I(x,y) is denoted by each of the six original cropped gray scale images, $P_{I(x,y)}(v)$ is the histogram representing probability of an occurrence of a pixel of gray level $v$, $n_v$ is the number of pixels having each pixel value is $v$ and $N$(width×height) is the total number of pixels, and $CH_{I(x,y)}(v)$ is the cumulative histogram(CH) function up to the gray level $v$ for an image I(x,y)[16],[17], where $0 \leq v \leq 255$. The CH ($v$) is measured by summing up the all histogram values from gray level 0 to $v$ [20]. The new filtering image, $I_{FI}(x,y)$ is achieved when CH value is not exceeded the threshold value $Th$ and the $I_{FI}(x, y)$ image only contains the white pixels of our specific desired connected component area. Figure 3(e) is shown the respective white pixel's connected component of all filtering images for right-eyebrow, right-eye, left-eyebrow, left-eye, nose, and mouth region. Three different groups of threshold values are used for our evaluation purpose. One for eyebrows region($0.01 \leq Th \leq 0.25$) another for eyes and mouth regions ($0.01 \leq Th \leq 0.10$) and the other for nose region ($0.001 \leq Th \leq 0.010$) because nostrils contain minimum numbers of low intensity pixels of original image compare to eyebrows, eyes and mouth region (see Figure 5) [4].

### 3.1 Eyebrow, Eye and Mouth Corner Points Detection

A simple linear search concept is applied on right-eyebrow, right-eye, left-eyebrow left-eye, and mouth filtering images to detect the first white pixel locations as the candidate points:
(1) Starting from top-left position for right corner points and (2) starting from top-right position for left corner points to search downward direction for eyebrows corner points
(3) Starting from bottom-left position for right corner points and (4) starting from bottom-right position for left corner points to search upward direction for eyes and mouth corner points.
The located first white pixel's positions are the candidate corner points.

### 3.2 Nostrils Detection and Nose Tip Calculation

A contour algorithm, using connected component, is applied on nose filtering image to select the last (right nostril) and the previous last (left nostril) contours from bottom to upward direction. Then the last and the previous last contour's element locations are sorted as an ascending order according to horizontal direction(x-value). The locations of the last element (right nostril point) of the last contour and the first element (left nostril point) of the previous last contour are the candidate nostrils. Nose tip is computed as the mid point between nostrils because the nose tip conveys the highest gray scale value so that nose filtering image shows insufficient information about it(see the 2nd filtering image started from the lower position of Figure 3(e))[6], [18].

All of the detected twelve corner points are indicated as 'black plus symbols', and only calculated nose tip is indicated as 'black solid circle' as shown in Figure 6.

### 3.3 Proposed Algorithm

The proposed algorithm is organized by three sections, which are included "preprocessing", "processing", and "detection" sections (see Figure 1). The preprocessing section detects the face and its location and then crops the face, right-eyebrow, right-eye, left-eyebrow, left-eye, nose, and mouth regions in an image. We assume that as a frontal face image, the eyebrows & eyes, nose, and mouth are located in upper half, middle and lower parts, respectively, in an image (see Figure 2 and Figure 3). In the processing section, the cropped images i.e. six ROIs such as right-eyebrow, right-eye, left-eyebrow, left-eye, nose, and mouth are converted into filtering images by applying CH method (using equations (1),(2), & (3)) [16],[17]. Applying simple linear search and contour concepts on these filtering images, the detection section finds out the all facial feature points such as right & left eyebrows corners, right & left eyes corners, nostrils, and mouth corners. The step by step procedures of our proposed algorithm are described as follows.

### A. Preprocessing section

1. Input: $I_{whole-face-window}(x,y)$ =Frontal face gray scale image having head and shoulder (whole face window)(see Figure 3(a) ).
2. Detect and localize the face by applying the OpenCV face detection algorithm [19].
3. Detect the regions of interest (ROI) for face, right-eyebrow, right-eye, left-eyebrow, left-eye, nose, and mouth by applying the OpenCV ROI library functions [19] and then we build the following new images.

   (**a**) $I_{face}(x,y)$ =New image having only face area and its size is W×H(see Figure 2 and Figure 3(b)) Where, W=image width, H=image height.
   (**b**) $I_{eyebrow-right}(x,y)$ =New image having only right eyebrow area and its size is 0.375W×0.12H(See Figure 2 and Figure 3(d)).
   (**c**) $I_{eyebrow-left}(x,y)$ =New image having only left eyebrow area and its size is 0.375W×0.12H(See Figure 2 and Figure 3(d)).
   (**d**) $I_{eye-right}(x,y)$ =New image having only right eye area and its size is 0.375W×0.25H(See Figure 2 and Figure 3(d)).
   (**e**) $I_{eye-left}(x,y)$ =New image having only left eye area and its size is 0.375W×0.25H(See Figure 2 and Figure 3(d)).
   (**f**) $I_{nose}(x,y)$ =New image having only nose area and its size is 0.50W×0.19H(See Figure 2 and Figure 3(d)).
   (**g**) $I_{mouth}(x,y)$ =New image having only mouth area and its size is 0.50W×0.16H(See Figure 2 and Figure 3(d)).

### B. Processing section

4. Apply CH method(using equations (1),(2), & (3)) [16],[17] on the above six ROIs such as $I_{eyebrow-right}(x,y)$, $I_{eyebrow-left}(x,y)$, $I_{eye-right}(x,y)$, $I_{eye-left}(x,y)$, $I_{nose}(x,y)$, and $I_{mouth}(x,y)$ images(see Figure 3(d)) and convert it into new filtering(binary) images such as $I_{FI\_eyebrow-right}(x,y)$, $I_{FI\_eyebrow-left}(x,y)$, $I_{FI\_eye-right}(x,y)$, $I_{FI\_eye-left}(x,y)$, $I_{FI\_nose}(x,y)$, and $I_{FI\_mouth}(x,y)$ for different threshold values(see Figure 3(e)).

### C. Detection section

5. (**a**) A simple linear search concept is applied on filtering images such as $I_{FI\_eyebrow-right}(x,y)$, $I_{FI\_eyebrow-left}(x,y)$, $I_{FI\_eye-right}(x,y)$, $I_{FI\_eye-left}(x,y)$, and $I_{FI\_mouth}(x,y)$ for eyebrows, eyes and mouth corner points, then find out the first white pixel location as top-down approach on eyebrows filtering images and bottom-up approach on eyes & mouth filtering images. To locate for all corner points :
(1) Starting searches from top-left position for right corner points and (2) starting searches from top-right position for left corner points to search top-down approach for eyebrows corner points.
(3) Starting searches from bottom-left position for right corner points and (4) starting searches from bottom-right position for left corner points to search bottom-up approach for eyes & mouth corner points.

   (**b**) Apply the OpenCV contour library function on filtering image, $I_{FI\_nose}(x,y)$ for nostrils; then consider the locations of the last element(right nostril point) and the first element (left nostril point) for the last and the previous last contours as a bottom-up approach where, the contour's element locations are sorted horizontally (x-value) as an ascending order[19]. The sorted locations of contour elements for the last (Contour P) and the previous last (Contour Q) contours are shown in figure 4.

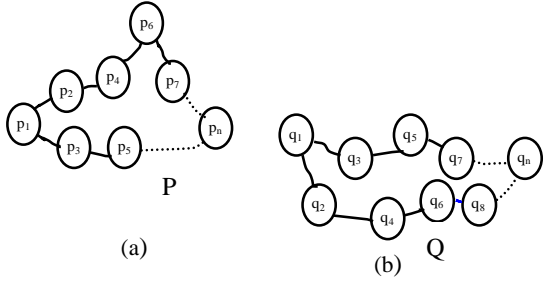

Figure 4. The locations of contours elements of Nose filtering image :(a) Sorted locations(x-value) of the last contour elements≡P($p_1$, $p_2$,……,$p_n$), (b) Sorted locations(x-value) of the previous last contour elements≡Q($q_1$, $q_2$,……,$q_n$).

**The locations of the last contour elements are:**
$p_1(x_{11}, y_{11})$, $p_2(x_{12}, y_{12})$,……,$p_n(x_{1n}, y_{1n})$. Where, $x_{11} < x_{12}……< x_{1n}$.
**Right Nostril**=The last element of the last contour
=$\mathbf{p_n(x_{1n}, y_{1n})}$

**The locations of the previous last contour elements are:**
$q_1(x_{21}, y_{21})$, $q_2(x_{22}, y_{22})$,……,$q_n(x_{2n}, y_{2n})$. Where, $x_{21} < x_{22}……< x_{2n}$.
**Left Nostril**=The first element of the previous last contour=$\mathbf{q_1(x_{21}, y_{21})}$

(**c**) The mid point (x-value) between nostrils and minimum y-value for nose tip is:
$X_{nose\ tip}$=Integer value of $(x_{1n}+x_{21})/2$
$Y_{nose\ tip}=((y_{1n}<y_{21})?y_{1n} :y_{21})-8$
**So Nose Tip ≡( $X_{nose\ tip}$ ,$Y_{nose\ tip}$ )**

At last, the detected points are transferred to the $I_{face}(x,y)$ image (see Figure 3(b) and Figure 6).

## 4 Experimental Results

### 4.1 Face Database

The work described in this paper is used head-shoulder BioID face database [15].The dataset with the frontal view of a face of one out of 23 different test persons consists of 1521 gray level images having properties of different illumination, face area, complex background with a resolution of 384×286 pixel. During evaluation, some images are omitted due to :(1) detecting false region (not face) by Viola-Jones face detector [14] and (2) person with large size eye glasses and highly dense moustache or beard as a complex background property of an image.

### 4.2 Results

The proposed algorithm was primarily developed and tested on Code::Blocks the open source, cross-platform combine with c++ language, and GNU GCC compiler. Some OpenCV library functions were used for face detection and localization, cropping and also connected component (contour algorithm) purpose [19]. During evaluation, three different groups of threshold values were used for our CH analysis (using equations (1), (2), & (3)) [16], [17]. One is $0.01 \leq Th \leq 0.25$ for locating eyebrows corner points, another is $0.01 \leq Th \leq 0.10$ for locating eyes and mouth corner points and the other is $0.001 \leq Th \leq 0.010$ for locating nostrils. Figure 5 shows the detection rate of twelve corner points by using different threshold values. Figure 5(a) shows single corner, both corners and overall detection rate for right eyebrow, left eyebrow corner points, Figure 5(b) shows single nostril, both nostrils and overall detection rate for nostrils and Figure 5(c) shows single corner, both corners and overall detection rate for right eye, left eye, and mouth corner points. The combination of single corner and both corners detection rate is considered as the overall detection rate. Threshold values 0.220, 0.240, 0.070, 0.060, 0.004, and 0.060 produce the detection rates 92.56%, 96.83% 97.92%, 98.02%, 89.58%, and 96.73% for right-eyebrow corners, left-eyebrow corners, right-eye corners, left-eye corners, nostrils, and mouth corners, respectively. Table 1 indicates the results of our facial feature extraction algorithm, where the overall an average detection rate is 95.27%. We compared our algorithm with R.S. Feris, et al. [6] and D. Vukadinovic, M. Pantic [2]. The comparison results are shown in table 2. Some of the detection results are shown in the Figure 6.

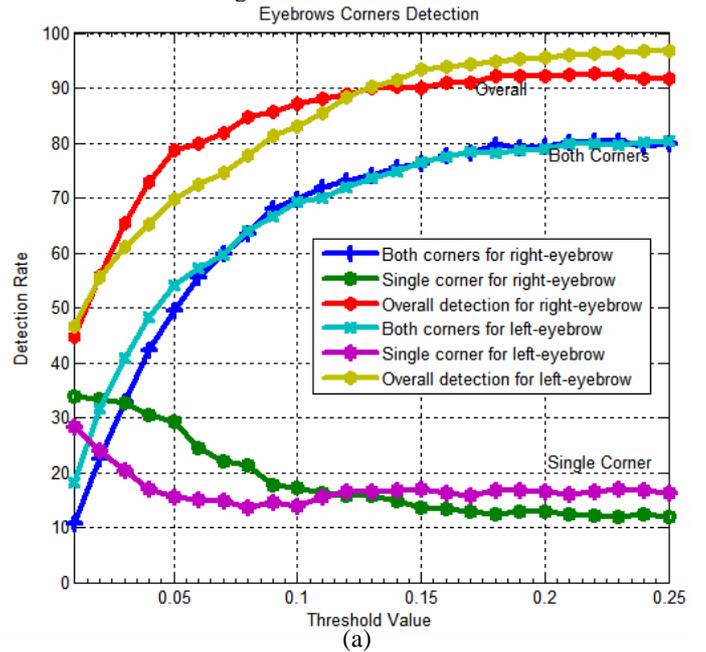

(a)

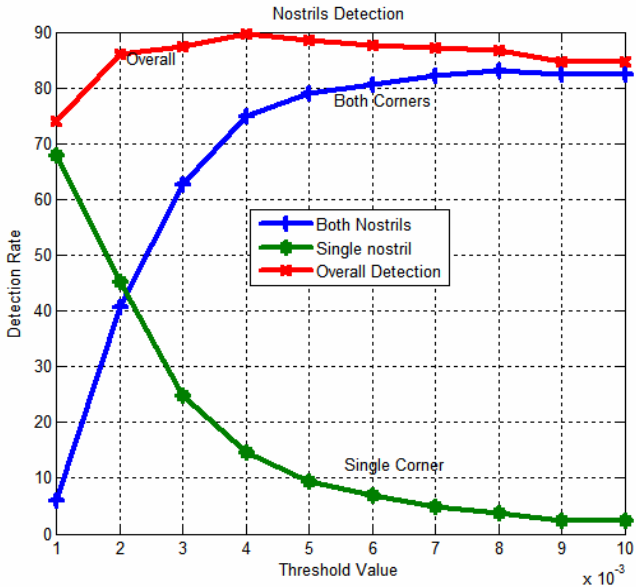

(b)

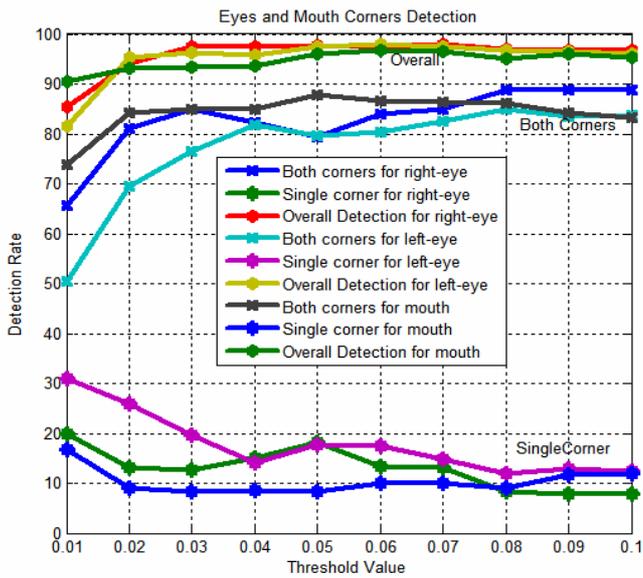

(c)

Figure 5. Detection Rate using different threshold values of CH method on BioID face database: (a) Eyebrows Detection Curves (Single, Both, Overall), (b) Nostrils Detection Curves (Single, Both, Overall), (c) Eyes and Mouth Corners Detection Curves (Single, Both, Overall).

Using the six relevant regions of a frontal view face image such as right eyebrow, left eyebrow, right eye, left eye, nose, and mouth areas are shown an average detection rate is 95.27% (See the table 1), whereas using the four relevant regions such as right eye, left eye, nose, and mouth areas are shown an average detection rate is 95.56% [15], [20].

Table 1: Table of feature points detection rate

| Features | Detection Rate (%) for both Points/ Corners | Detection Rate (%) for single Point/ Corner | Overall Detection Rate (%) | Threshold Value for CH Method |
|---|---|---|---|---|
| Right Eyebrow | 80.36 | 12.20 | 92.56 | 0.220 |
| Left Eyebrow | 80.16 | 16.67 | 96.83 | 0.240 |
| Right Eye | 84.82 | 13.10 | 97.92 | 0.070 |
| Left Eye | 80.46 | 17.56 | 98.02 | 0.060 |
| Nostrils | 75.00 | 14.58 | 89.58 | 0.004 |
| Mouth Corners | 86.71 | 10.02 | 96.73 | 0.060 |
| **Average** | **81.25** | **14.02** | **95.27** | - |

Table 2: Comparisons with 2-level GWN [6] and GFBBC [2]

| Algorithms | Average Detection rate (%) |
|---|---|
| 2-level GWN | 92.87 |
| GFBBC | 93.00 |
| Ours | 95.27 |

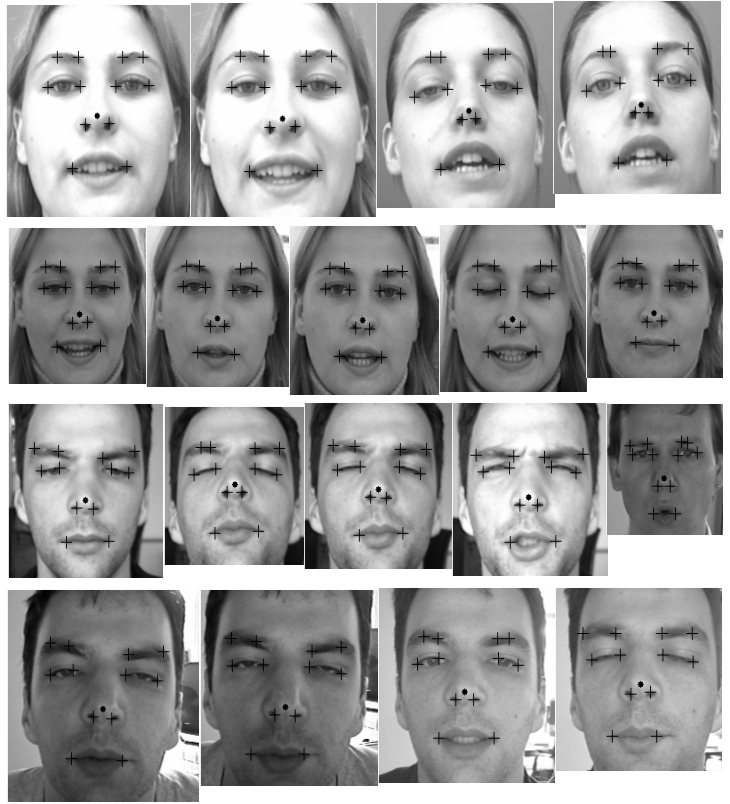

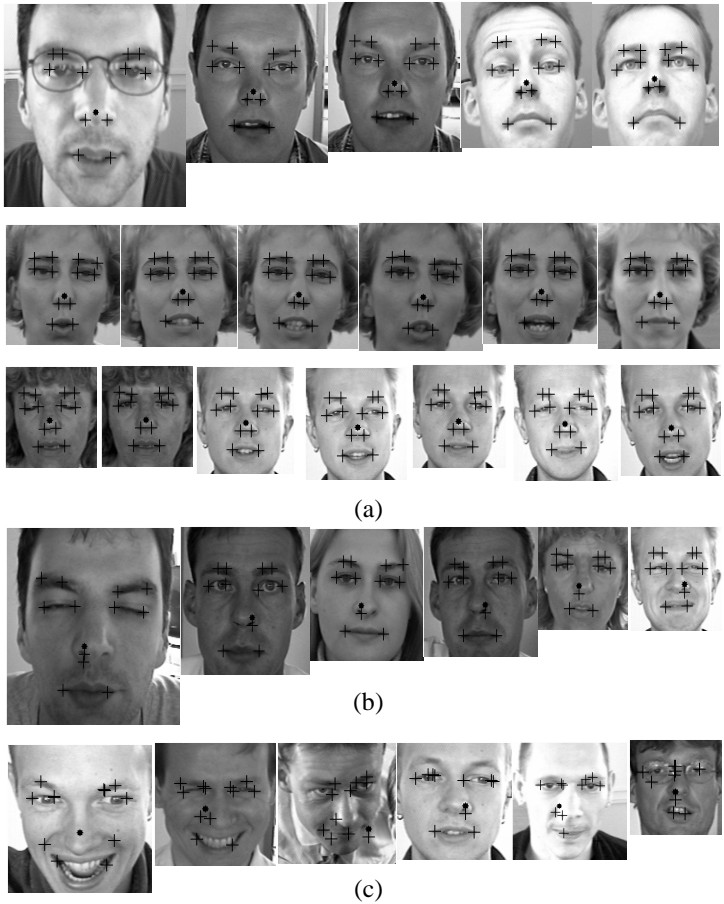

(a)

(b)

(c)

Figure 6. Result of detected feature points:(a) Some true detection, (b) Some single nostril detection and (c) Some false detection.

## 5 Conclusions and Future Work

In this paper, we have shown how salient facial features are extracted based on cumulative histogram CH method in an adaptive manner combined with face detector, simple linear search, and also connected component concepts i.e. contour algorithm in various expression and illumination conditions in an image. Image segments are converted into filtering images with the help of CH approach by varying different threshold values instead of applying morphological operations. Our algorithm was assessed on free accessible BioID gray scale frontal face database. The experimental results confirmed the higher detection rate as compare to other well known facial feature extraction algorithms.

Future work will concentrate to improve the detection rate of both corner points instead of single corner point by using a single threshold group instead of multiple threshold groups and face recognition, as well.

## References

[1] W. Zhao, R. Chellappa, P. J. Phillips, and A. Rosenfeld, "Face Recognition: A Literature Survey", *ACM Computing Surveys*, Vol. 35, No. 4, December 2003.

[2] D. Vukadinovic and M. Pantic," Fully Automatic Facial Feature Point Detection Using Gabor Feature Based Boosted Classifiers", *2005 IEEE International Conference on Systems, Man and Cybernetics Waikoloa*, Hawaii, October 10-12, 2005.

[3] http://eprints.um.edu.my/877/1/GS10-4.pdf

[4] Wei Jen Chew, Kah Phooi Seng, and Li-Minn Ang, "Nose Tip Detection on a Three-Dimensional Face Range Image Invariant to Head Pose", *Proceedings of The International Multi Conference of Engineers and Computer Scientists 2009*, Vol I, IMECS 2009, March 18-20, 2009, Hong Kong.

[5] I. Matthews and S. Baker, "Active Appearance Models Revisited", *Int'l Journal Computer Vision*, vol. 60, no. 2, pp.135-164, 2004.

[6] R.S. Feris, et al., "Hierarchical Wavelet Networks for Facial Feature Localization", *Proc. IEEE Int'l Conf. Face and Gesture Recognition*, pp. 118-123, 2002.

[7] E. Holden, R. Owens, "Automatic Facial Point Detection", *Proc. The 5th Asian Conf. on Computer Vision, 23-25 January 2002, Melbourne, Australia*.

[8] M. J. T. Reinders, et al., "Locating Facial Features in Image Sequences using Neural Networks", *Proc. IEEE Int'l Conf. Face and Gesture Recognition,* pp230-235, 1996.

[9] C. Hu, et al.,"Real-time view-based face alignment using active wavelet networks", *Proc. IEEE Int'l Workshop Analysis and Modeling of Faces and Gestures,* pp. 215-221, 2003.

[10] S. Yan, et al., "Face Alignment using View-Based Direct Appearance Models", *Int'l J. Imaging Systems and Technology*, vol. 13, no. 1, pp. 106-112, 2003.

[11] L. Wiskott, et al., "Face Recognition by Elastic Bunch Graph. Matching" *IEEE Trans. Pattern Analysis and Machine Intelligence*, vol. 19, no.7, pp. 775-779, 1979.

[12] D. Cristinacce, T. Cootes, "Facial Feature Detection Using AdaBoost with Shape Constrains", *British Machine Vision Conference*, 2003.

[13] L. Chen, et al., "3D Shape Constraint for Facial Feature Localization using Probabilistic-like Output", *Proc. IEEE Int'l Workshop Analysis and Modeling of Faces and Gestures*, pp. 302-307, 2004.

[14] P. Viola and M. J. Jones, "Robust Real-time Object Detection", *International Journal of Computer Vision*, Vol. 57, No.2, p.137-154, 2004.

[15] BioID Face Database, Available: http://www.bioid.com/downloads/facedb/index.php

[16] Joung-Youn Kim, Lee-Sup Kim, and Seung-Ho Hwang, "An Advanced Contrast Enhancement Using Partially Overlapped Sub-Block Histogram", *IEEE Transactions On*

[17] Mansour Asadifard, and Jamshid Shanbezadeh, "Automatic Adaptive Center of Pupil Detection Using Face Detection and CDF Analysis", *Proceedings of The International Multi Conference of Engineers and Computer Scientists 2010*, pp130-133, IMECS 2010, 17-19 March, 2010, Hong Kong.

[18] S. Jahanbin, et al., "Automated Facial Feature Detection from Portrait and Range Images", *SSIAI '08 Proc. Of the 2008 IEEE Southwest Symposium on Image Analysis and Interpretation*, 24-26 March 2008.

[19] http://sourceforge.net/projects/opencvlibrary/files/opencv-win/2.0/OpenCV-2.0.0a-win32.exe/download

[20] Sushil Kumar Paul, Saida Bouakaz and Mohammad Shorif Uddin, "Automatic Adaptive Facial Feature Extraction using CDF Analysis", *Proceedings of The International Conference on Digital Information and Communication Technology and its Applications,* Communications in Computer and Information Science, Volume 166, Part 1, Pages 327-338, DICTAP2011, 21-23 June 2011, Dijon, France.
Circuits And Systems For Video Technology*, Vol. 11, No. 4, April 2001.
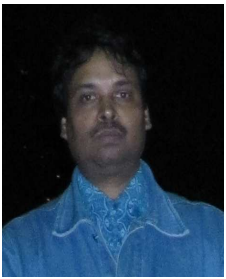

**Sushil Kumar Paul** is currently pursuing his PhD in Face Recognition, Department of Computer Science and Engineering (CSE) from Jahangirnagar University (JU), Dhaka. He has completed his Master of Science in CSE from United International University (UIU) and Bachelor of Science in Electrical and Electronic Engineering from Bangladesh University of Engineering and Technology (BUET), Dhaka. He has worked as an eLINK(East-West Link for Innovation, Networking and Knowledge Exchange) scholar in LIRIS Laboratory, University Claude Bernard Lyon1, France in 2010. He has twelve years of teaching experience as an instructor (Computer/Electronics) in many Polytechnic Institutes under Directorate of Technical Education (DTE) in Bangladesh. He has also been participated many short terms foreign & local training programs and he is currently working as Principal, Narsingdi Polytechnic Institute under DTE. He has published one international conference paper and also worked as a reviewer. He is a writer of one local textbook and a head examiner under Bangladesh Technical Education Board (BTEB). His research interests are in computer vision, digital imaging, machine learning, and data mining including face, facial expression and motion recognition. He is also a member of IEB, BCS and BUET87 foundation.

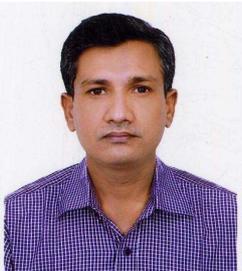

**Mohammad Shorif Uddin** received his PhD in Information Science from Kyoto Institute of Technology, Japan, Masters of Education in Technology Education from Shiga University, Japan and Bachelor of Science in Electrical and Electronic Engineering from Bangladesh University of Engineering and Technology (BUET). He joined the Department of Computer Science and Engineering, Jahangirnagar University, Dhaka in 1992 and currently serves as a Professor of this department. He began his teaching career in 1991 as a Lecturer of the Department of Electrical and Electronic Engineering, Chittagong University of Engineering and Technology (CUET). He undertook postdoctoral research at Bioinformatics Institute, A-STAR, Singapore, Toyota Technological Institute, Japan and Kyoto Institute of Technology, Japan. His research is motivated by applications in the fields of computer vision, pattern recognition, blind navigation, bio-imaging, medical diagnosis and disaster prevention. He has published a remarkable number of papers in peer-reviewed international journals and conference proceedings. He holds two patents for his scientific inventions. He received the Best Presenter Award from the International Conference on Computer Vision and Graphics (ICCVG 2004), Warsaw, Poland. He is the co-author of two books. He is also a member of IEEE, SPIE, IEB and a senior member of IACSIT.

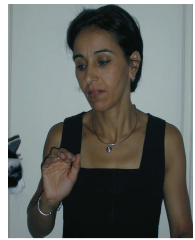

**Saida Bouakaz** received her PhD from Joseph Fourier University in Grenoble, France. She is currently a professor in the Computer Science and head of the SAARA Research Team, LIRIS Lab at Claude Bernard University, Lyon1. Her research interests are in computer vision and graphics including motion capture and analysis, face, facial expression and gesture recognition, facial animation.